\documentclass{article}
\usepackage{kr}

\usepackage{times}
\usepackage{soul}
\usepackage{url}
\usepackage[hidelinks]{hyperref}
\usepackage[utf8]{inputenc}
\usepackage[small]{caption}
\usepackage{graphicx}
\usepackage{amsmath}
\usepackage{amsthm}
\usepackage{booktabs}
\usepackage{algorithm}
\usepackage{algorithmic}
\usepackage{wrapfig}
\usepackage{multirow}
\usepackage{subcaption}
\usepackage{graphicx}
\usepackage{xcolor}

\usepackage{amsfonts}

\usepackage{natbib}

\usepackage{amssymb}
\usepackage{tikz}
\usetikzlibrary{intersections}

\newtheorem{example}{Example}
\newtheorem{theorem}{Theorem}
\newtheorem{lemma}{Lemma}
\newtheorem{definition}{Definition}

\newcommand\algorithmicprocedure{\textbf{procedure}}
\newcommand{\algorithmicendprocedure}{\algorithmicend\ \algorithmicprocedure}
\makeatletter
\newcommand\PROCEDURE[3][default]{%
  \ALC@it
  \algorithmicprocedure\ \textsc{#2}(#3)%
  \ALC@com{#1}%
  \begin{ALC@prc}%
}
\newcommand\ENDPROCEDURE{%
  \end{ALC@prc}%
  \ifthenelse{\boolean{ALC@noend}}{}{%
    \ALC@it\algorithmicendprocedure
  }%
}

\newcommand{\alglinelabel}{%
  \addtocounter{ALC@line}{-1}%
  \refstepcounter{ALC@line}%
  \label%
}

\def\C2{\mathrm{C}^2}

\def\1{\mathbf{1}}

\definecolor{color1}{HTML}{D81B60}
\definecolor{color2}{HTML}{1E88E5}
\definecolor{color4}{HTML}{FFC107}%
\definecolor{color3}{HTML}{004D40}

\title{Logical Distillation of Graph Neural Networks}

\author{%
Alexander Pluska, Pascal Welke, Thomas G{\"a}rtner, and Sagar Malhotra\\
\affiliations
TU Wien\\
\emails
  \{alexander.pluska, pascal.welke, thomas.gaertner, sagar.malhotra\}@tuwien.ac.at
}

\begin{document}

\maketitle

\def\KR{}

\begin{abstract}
 We present a logic based interpretable model for learning on graphs and an algorithm to distill this model from a Graph Neural Network (GNN). 
    Recent results have shown connections between the expressivity of GNNs and the two-variable fragment of first-order logic with counting quantifiers ($\C2$). 
    We introduce a decision-tree based model which leverages an extension of $\C2$ to distill interpretable logical classifiers from GNNs. 
    We test our approach on multiple GNN architectures. 
    The distilled models are interpretable, succinct, and attain similar accuracy to the underlying GNN. 
    Furthermore, when the ground truth is expressible in $\C2$, our approach outperforms the  GNN.

\end{abstract}

\section{Introduction}

We present and evaluate an algorithm for distilling \emph{Graph Neural Networks} (GNNs) into a symbolic model. Our distillation algorithm relies on a novel model called \emph{Iterated Decision Tree} (IDT), which is tailored to represent logical formulas represented by GNNs.
GNNs play a crucial role in safety-critical applications like drug discovery and in cost-critical applications like large-scale transport routing. However, most GNN models are black-box in nature and their internal representations are opaque to any human or computer-aided formal scrutiny. Hence, interpreting and explaining GNN predictions is a fundamental problem of significant research interest. Although many results have analyzed the expressivity of GNNs in terms of formal languages like first-order logic, extracting the logical classifiers expressed by GNNs remains largely unexplored. We aim to fill this gap by developing a distillation model aimed at extracting logical classifiers expressed by GNNs.

The key motivation for our model is the close relationship between GNNs and first-order logic with only two variables and counting quantifiers $\C2$ \citep{Barcel2019LogicalEO, grohe2021logic, Grohe2023TheDC}.
Hence our model, the IDT, is designed to express any $\C2$ formula. An IDT consists of a sequence of decision trees. Each decision tree expresses a number of unary $\C2$ formulas of quantifier depth one. Combining multiple such decision trees enables us to express formulas of larger quantifier depth. 
Additionally, we propose an extension of $\C2$ that can capture operations like mean aggregation, which are common in GNNs, and incorporate it into IDTs. 
Our distillation algorithm is able to exploit intermediate node representations from each message-passing layer of a GNN to iteratively learn decision trees of an IDT. Although the learning process for IDTs is guided by the GNN, our empirical results show that the logic-based inductive bias incentivizes succinct and interpretable models.

We test IDTs on multiple synthetic and real-world datasets, performing distillation on two prominent GNN architectures: Graph Isomorphism Networks (GIN) \citep{xu19howpowerful} and Graph Convolution Networks (GCN) \citep{kipf2016semi}. Our algorithm consistently distills IDTs that are succinct and have comparable predictive performance to the underlying GNN. Furthermore, when the ground truth is a $\C2$ formula, the distilled IDT exhibits better generalization, outperforming the GNN on the test data. Qualitatively, we find that our method can provide new insights. %
For instance, on the AIDS dataset~\citep{riesen2008iam}, IDTs infer a very simple high-performing rule that achieves over 99\% classification accuracy. This rule classifies graphs based on their number of nodes being smaller or larger than 12. To the best of our knowledge, none of the existing GNNs or explanation methods have been able to infer this rule.

In the next section we discuss the relevant related work. In Section \ref{Sec:Preliminaries} we discuss the necessary background on graphs, logic and GNNs. We present IDTs in Section~\ref{Sec:IDT} and show how IDTs can be distilled from GNNs in Section~\ref{Sec:Training}. 
In Section~\ref{sec:cp}, we introduce an extension of $\C2$, which allows us to learn more expressive IDTs. 
We empirically evaluate IDTs in Section~\ref{Sec:Experiments} and analyze some of the obtained logical explanations in Section~\ref{Sec:Explainability}. 
Finally, we summarize our work and discuss future research directions in  Section~\ref{Sec:Conclusion}. %

\section{Related Work}
\label{sec:relatedwork}

Our work is related to explanation methods of GNNs \citep{Exp_Survey} and to logical approaches \citep{Barcel2019LogicalEO, grohe2021logic, Grohe2023TheDC} for analyzing their expressivity.
Explanation methods aim to derive insights about the process underlying the \emph{model predictions}. Although IDTs may aid such understanding, our goal is different. We aim to distill an interpretable classification model for the \emph{data}, while using trained GNN model as guidance for the learning process. Hence, we want our model to not only be interpretable, but also to generalize well. In the literature, methods that provide a global explainer, i.e., an interpretable surrogate model \citep{azzolin2022global}, can be adapted to yield classification models for the underlying data. However, such models come at a significant cost to the accuracy. 
 
Our work is also loosely connected to the general problem of learning desision trees from neural networks. This problem has already been extensively investigated for tabular data \citep{DBLP:conf/nips/CravenS95,DBLP:journals/pr/KrishnanSB99,DBLP:conf/kdd/Boz02,DBLP:conf/flairs/DanceyMB04,DBLP:journals/ai/SetzuGMTPG21}. Furthermore, recent works have also investigated the tweaking of learning process or the  neural architecture itself for learning decision trees \citep{DBLP:conf/icmla/SchaafHM19,DBLP:conf/aaai/WuHPZ0D18,DBLP:journals/corr/abs-1806-06988,DBLP:conf/ijcai/KontschiederFCB16}. Although these results are related to our approach in spirit, our work is fundamentally different in its theoretical motivation and the learning procedure. GNNs expressivity is deeply connected to that of first-order logic \citep{Barcel2019LogicalEO,grohe2021logic,Grohe2023TheDC}. Hence, using logic-based decision trees is a natural choice for learning decision trees from GNNs. 

Explanation methods that distill surrogate models from GNNs come closest to our approach. 
\citet{azzolin2022global} first derive instance-level local subgraphs as explanations and then cluster them to extrapolate a model-level Boolean formula using the subgraphs as concepts.
\citet{yuan2020xgnn} base their approach on input-optimization, i.e. globally reducing graphs to a number of instances for which the explanation is then given.
Their approach requires prior domain knowledge. \citet{graphchef2024} first compute (almost) categorical layer wise node representations using GNN layers with Gumbel-Softmax update functions. 
Subsequently, they replace the neural networks by decision trees trained on the categorical node representations.
Their approach results in an interpretable message passing scheme based on intermediate categorical node states but requires to train a specific GNN architecture. All three approaches use graphs, sub-graphs, or their combinations as the explanation model. This restricts these methods, as many simple and important constraints, e.g. a graph has more than 12 nodes, can not easily be expressed in terms of subgraphs. Similarly to us, \citet{kohler2024descriptionlogics} use a description logic based method for explaining GNNs. Their method uses GNNs as a black box, whereas IDTs are distilled layer-by-layer, exploiting GNNs representations at each iteration. Furthermore, their approach currently does not support counting quantifiers.

\section{Background}\label{Sec:Preliminaries}

We use $[n]$ to denote the set $\{0, \dots, n-1\}$. For a matrix $A$ we write $A_{ij}$ for the entry in the $i$-th row and $j$-th column. $A^{T}$ is the transpose of $A$ and $A^{-1}$ its inverse. We write $I$ for the identity matrix, i.e., a matrix with all diagonal entries equal to one and all non-diagonal entries equal to zero. We write $1$ to denote the matrix with all entries equal to one. 

A \emph{simple undirected graph} $G$ consists of a set of nodes $V$ and a set of edges $E$. Without loss of generality, a finite set of nodes $V$ is given as $V = \{v_i\}_{i\in [|V|]}$. The adjacency matrix $A$ of a graph is a symmetric matrix, where $A_{ij} = 1$ if there is an edge connecting $v_i$ and $v_j$, otherwise $A_{ij} = 0$. We use $N(v)$ to denote the neighbors of a node $v$ and $d_v$ to denote its degree. In this paper, graphs are always simple and undirected.
We discuss possible generalizations in Section~\ref{Sec:Conclusion}.

A \emph{tree} is a connected acyclic graph. A \emph{rooted tree} is a tree with a designated root node. The \emph{depth} of a node $v$ in a rooted tree is the length of the unique path from the root to $v$. The \emph{depth} of a rooted tree is defined as the maximum depth of its nodes. Each node in a rooted tree, except the root node, has a unique \emph{parent node}, which is the only adjacent node with a smaller depth. The \emph{children} of a node are the nodes adjacent to it with a larger depth. A \emph{leaf node} is a node without children. A \emph{binary tree} is a rooted tree in which each node has at most two children. A binary tree of a given depth $h$ is perfect if it has $2^{h}$ leaves of depth $h$.

\subsection{Graph Neural Networks}
\label{subsec:gnn}

\emph{Graph neural networks} (GNNs) are a class of deep learning models that operate on graphs. We will consider message passing GNNs consisting of a sequence of layers that iteratively combine the feature vector of every node with the multiset of feature vectors of its neighbors. Formally, let $\mathsf{agg}_k$ and $\mathsf{comb}_k$ for $k \in [l]$ be \emph{aggregation} and \emph{combination} functions. We assume that each node has an associated initial Boolean feature vector $x_v = x_v^{(0)}$. A GNN computes a vector $x_{v}^{(k)}$ for every node $v$ via the following recursive formula
\begin{equation} \label{eq: GNN_Recursion}
x_v^{(k+1)} = \mathsf{comb}_{k+1}(x_v^{(k)}, \mathsf{agg}_{k+1}(\{\!\!\{x_w^{(k)}: w\in N(v)\}\!\!\})),
\end{equation}
where $k \in [l]$. The vectors $x_{v}^{(l)}$ are then \emph{pooled}  
\begin{equation}
\hat y = \mathsf{pool}(\{\!\!\{x_v^{(l)} : v\in V\}\!\!\})
\end{equation}
to give a single graph vector $\hat{y}$, the output of the GNN. Note that in our theoretical framework, $\mathsf{pool}$ is not limited to common pooling operations such as mean \ifdefined\KR or sum \fi but can also include more complicated operations, e.g., a neural network.

\begin{example}\label{ex:gcn}
A special case of the above architecture pattern is the GCN \citep{kipf2016semi}, in which 
\ifdefined\ICML\vspace*{-2mm}\fi
\[\mathsf{agg}(\{\!\!\{x_w^{(k)}:w\in N(v)\}\!\!\}) = \sum_{w\in N(v)}\frac{x_w^{(k)}}{\sqrt{d_w}}\]
\ifdefined\ICML\vspace*{-2mm}\fi
\[\mathsf{comb}(x_v^{(k)}, a) = \sigma\left(W^{(k)}\left(\frac{x_v^{(k)}}{d_v} + \frac{a}{\sqrt d_v}\right)\right)\]
\ifdefined\ICML\vspace*{-2mm}\fi
\[\mathsf{pool}(\{\!\!\{x_v^{(l)}\}\!\!\}) = \mathsf{MLP}\left(\sigma'\left(\frac{1}{|V|}\sum_{v\in V}x_v^{(l)}\right)\right)\]
where: $W^{(k)}$ are learned weight matrices; $\sigma$ and  $\sigma'$ are non-linear activation functions and $\mathsf{MLP}$ is a function computed by a Multilayer Perceptron.
\end{example}

\subsection{Graphs and Logic}
\label{subsec: graph_logic}
$\C2$ is the fragment of first-order logic with only two variables $v, w$. Besides the usual existential ($\exists$) and universal ($\forall$) quantifiers, $\C2$ also admits counting quantifiers of the form $\exists^{\geq n}, \exists^{\leq n}$ and $\exists^{=n}$, which stand for \emph{exist at least $n$, exist at most $n$} and \emph{exist exactly $n$} \citep{CaiFI89}. All $\C2$ sentences can be expressed in first-order logic without counting quantifiers albeit using more than two variables \citep{Grohe2023TheDC}. 
In this paper, we assume a first-order language on graphs consisting of exactly one binary predicate $E$ and $m$ unary predicates $\{U_j\}_{j\in [m]}$. $E(v,w)$ denotes that there is an edge between nodes $v$ and $w$. $U_{j}(v)$ denotes that the $j$-th node-attribute is true for the node $v$ in the graph. We use $U$ to represent a binary matrix with $|V|$ rows and $m$ columns. An entry $U_{ij}$ in $U$ is $1$ if $U_{j}(v_i)$ holds in a given graph. Hence, the matrix $U$ completely represents the interpretation of the atoms $\{U_j\}$ on a given graph. With a slight abuse of notation, we use $U_j$ also to denote the $j$-th column of $U$. Note that for a given graph $G$, its adjacency matrix $A$ completely encodes the interpretation of the predicate $E$.

\begin{example}\label{ex:graph}
  Consider the graph $G$ (shown below) with unary predicates $U_0, U_1$ where $U_0(v)$ is true if $v\in \{v_1, v_3\}$ and $U_1(v)$ holds if $v \in \{v_0, v_3\}$. Then we have

    \begin{center}

    \begin{tikzpicture}

    \node (g) {$G$};
    \node[anchor=west] (a) at (g.east) {
    \begin{tikzpicture}[x=0.8cm,y=0.8cm]
      \node (0) at (-2, 0) {$v_0$};
      \node (1) at (0, 0) {$v_1$};
      \node (2) at (-2, -2) {$v_2$};
      \node (3) at (0, -2) {$v_3$};
      \draw (0) -- (1);
      \draw (0) -- (2);
      \draw (1) -- (3);
      \draw (1) -- (2);
    \end{tikzpicture}
    };
    
    \node[anchor=west] (b) at (a.east) {
    $A = \begin{pmatrix}
    0 & 1 & 1 & 0\\
    1 & 0 & 1 & 1\\
    1 & 1 & 0 & 0\\
    0 & 1 & 0 & 0
  \end{pmatrix}
  $};
  
    \node[anchor=west] (c) at (b.east) {
    $U = \begin{pmatrix}
    0 & 1\\
    1 & 0\\
    0 & 0\\
    1 & 1
   \end{pmatrix}$
   };
\end{tikzpicture}

  \end{center}

Here, $U_0 = (0101)^{T}$ and $U_1 = (1001)^{T}$. The matrix product $AU_j$ gives us the number of neighbors of each node satisfying $U_j$. Similarly, $(1-A)U_j$ gives us all the non-neighbors of each node satisfying $U_j$, where $1$ is a matrix with all entries equal to $1$.
\end{example}

We will consider node and graph classifiers. For instance, the $\C2$ formula $\exists^{=2}y \ E(x,y)$ is a logical node classifier which characterizes nodes of a given graph with degree exactly 2. The formula has exactly one free variable $x$ and can, therefore, be evaluated on the nodes of a given graph. On the other hand, $\forall x \exists^{=2}y \ E(x,y)$ does not have any free variables. Therefore, it constitutes a graph classifier. It characterizes 2-regular graphs.
\citet{Barcel2019LogicalEO} and \citet{grohe2021logic} have shown multiple connections between $\C2$ and expressivity of GNNs. A key result is the following:

\begin{theorem}[\citet{Barcel2019LogicalEO}, Theorem 5.2]
\label{th:C2_GNN}
  Any classifier expressible in $\C2$ can be computed by a GNN.
\end{theorem}
Note that GNNs can also compute classifiers that are inexpressible in first-order logic.
Whether every first-order classifier computed by a GNN is expressible in $\C2$ is an open question. However, for GNNs without $\mathsf{pool}$ operation, first-order logic expressivity is completely characterized by the guarded fragment of $\C2$ \citep[Theorem 4.2]{Barcel2019LogicalEO}.  
The proof of Theorem~\ref{th:C2_GNN} builds on the equivalence between $\C2$ and the modal logic $\mathcal{EMLC}$ \citep{Modal_Lutz}. $\mathcal{EMLC}$  allows us to define a convenient grammar and serves as the motivation for our proposed model. We follow \citet[Appendix D]{Barcel2019LogicalEO} and introduce a language similar to their Lemma D.4.

\begin{definition}
\label{def: EMLC}
  A \emph{modal parameter} $S$ is one of the following
  \[0, 1, I, A, 1-I, 1-A, I + A, 1 - I - A.\]
  Given a graph $G$ and vertex $v$, the interpretation $\varepsilon_S(v)$ of $S$ on $v$ is defined as
  \begin{align*}
    \varepsilon_0(v) &:= \emptyset &
        \varepsilon_{1-I}(v) &:= V\setminus \{v\} \\
    \varepsilon_{1}(v) &:= V &
        \varepsilon_{1-A}(v) &:= V\setminus N(v) \\
    \varepsilon_I(v) &:= \{v\} &
        \varepsilon_{I+A}(v) &:= \{v\}\cup N(v)\\
    \varepsilon_A(v) &:= N(v) &
        \varepsilon_{1-I-A}(v) &:= V\setminus (\{v\}\cup N(v))
  \end{align*}
  An $\mathcal{EMLC}$ formula is then built by the following grammar:
  \[\varphi ::= U_j \mid \top\mid \varphi\wedge\varphi \mid \varphi\vee\varphi \mid\neg\varphi \mid S\varphi > n\]
  where $U_j$ ranges over all the unary predicates, $\top$ represents the predicate which is always true, S ranges over modal parameters, and $n$ ranges over $\mathbb N$. The semantics of the logical connectives ($\top$, $\land, \lor$, and $\neg$) are defined as usual. We say that $(G, v) \models S\varphi > n$ if  there are more than $n$ vertices $w\in\varepsilon_S(v)$ with $(G, w) \models\varphi$. We say $G\models \varphi$ if $(G, v)\models\varphi$ for all nodes $v\in G$. The depth of a formula is the maximal number of nested modal parameters.
\end{definition}

Definition~\ref{def: EMLC} can be used to express other comparison symbols, for instance one can define $S\varphi < n$ as $\neg(S\varphi > n-1)$ and $S\varphi = n$ as $\neg(S\varphi < n)\wedge \neg (S\varphi > n)$.
Computationally, $\mathcal{EMLC}$ formulas can be interpreted as matrix operations. The modal parameters are naturally interpreted as matrices, i.e., $0$ and $1$ as the square matrices with numerical entries zero and one everywhere, $I$ as the identity matrix, and $A$ as the adjacency matrix. 
Suppose $\varphi$ is given as a binary vector, i.e., the $i$-th entry of $\varphi$ is 1 if and only of $(G, v_i)\models \varphi$. Then the $i$-th entry of the matrix-vector product $S\varphi$ is the number of $w\in\varepsilon_S(v_i)$ such that $(G, w)\models\varphi$. %
Vectorizing $\wedge,\vee,\neg$ and $>$ gives a convenient method for determining which nodes satisfy a given $\mathcal{EMLC}$ formula.

\begin{example}\label{ex:EMLC}
  We have the following $\C2$\; formula $\varphi$ with a free variable $v$
\begin{equation}
\label{eq: example_eq}
    \varphi := \exists^{>1}w(E(v, w)\wedge\neg\exists^{=1}v(E(w, v)\wedge U_1(v))).
\end{equation}
\noindent
Hence, $\varphi$ is true for a node if it has more than one neighbor $w$ satisfying \[\neg\exists^{=1}v(E(w, v)\wedge U_1(v)),\]
that is, $w$ must not have exactly one neighbor satisfying $U_1$. Here, we have re-used the variable $v$. The inner-quantification is bound to the quantifier $\exists^{=1}$. Whereas the first occurance of $v$ is free. 
Note that $\varphi$ can be equivalently written as the following  $\mathcal{EMLC}$\; formula 
\begin{equation}
\label{eq: EMLC_eq}
    A(\neg(AU_1 = 1)) > 1.
\end{equation}
\noindent
When evaluating the formula \eqref{eq: EMLC_eq} on the graph given in Example~\ref{ex:graph},  we have %
  \begin{align*}
    A(\neg(AU_1 = 1)) > 1 &=
    A(\neg(A (1001)^T = 1)) > 1 \\ &=
    A(\neg((0210)^T = 1))) > 1\\ &=
    A(\neg((0010)^T)) > 1\\ &=
    A(1101)^T > 1\\ &=
    (1221)^T > 1\\ &=
    \left(0110\right)^T
    \end{align*}
\noindent
The second entry of $AU_1= A (1001)^T = (0210)^T$ represents
that $v_1$ has two neighbors $w$ for which $U_1(w)$ is true. Similarly, the third entry represents that  $v_2$ has exactly one such neighbor %
and the zero entries reflect that both $v_0$ and $v_3$ have no neighbors satisfying $U_1$. The subsequent computation shows that $v_1$ and $v_2$ satisfy $\varphi$ while $v_0$ and $v_3$ don't.
\end{example}

 The following result makes the connection between $\mathcal{EMLC}$ and $\C2$ explicit. A \emph{unary} $\C2$ formula is one in which, at most, one variable occurs freely.

\begin{theorem}[\citet{Barcel2019LogicalEO}, Theorem~D.3, Lemma~D.4]\label{thm:c2-emlc}
  For every $\mathcal{EMLC}$ formula there is an equivalent unary $\C2$ formula. Conversely, for every unary $\C2$ formula there is an equivalent $\mathcal{EMLC}$ formula.
\end{theorem}

\subsection{Decision Trees}

\begin{figure}
  \begin{center}
    \begin{tikzpicture}[level 1/.style={sibling distance=30mm, level distance=10mm},level 2/.style={sibling distance=10mm}]
      \node {$U_{1}$}
        child {
          node {$U_0$} {
            child {node {\color{color1}$\:v_2$} edge from parent node[above left, draw=none] {\tiny False}}
            child {node {\color{color1}$\:v_1$} edge from parent node[above right, draw=none] {\tiny True}}
          } edge from parent node[above left,draw=none] {\tiny False}
        }
        child {
          node {$\:U_{0}\:$} {
            child {node {$\:v_0$} edge from parent node[above left, draw=none] {\tiny False}}
            child {node {\color{color1}$\:v_3$} edge from parent node[above right, draw=none] {\tiny True}}
          }  edge from parent node[above right,draw=none] {\tiny True}
        };
    \end{tikzpicture}
    \caption{A simple decision tree. Each set of leaves can be interpreted as a formula, e.g. set of red leaves can be interpreted as $(\neg U_1\wedge \neg U_0) \vee(\neg U_1\wedge U_0)\vee(U_1\wedge U_0).$}
    \label{fig:dt}
    \vspace{-.2cm}
  \end{center}
\end{figure}
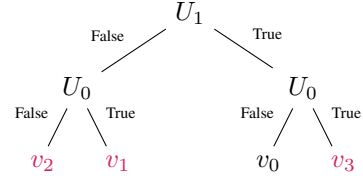

A \emph{decision tree} is a hierarchical, binary-tree structured, decision model. It assigns labels to samples through a sequence of binary decisions. Each leaf of a decision tree can be interpreted as a logical formula consisting of the conjunction of decisions from the root of the tree to the leaf. All samples satisfying this formula are assigned the label of the leaf. For example,  the leaf labeled $v_2$ in Figure~\ref{fig:dt} can be interpreted as $\neg U_1\wedge \neg U_0$, the leaf labeled $v_1$ as $\neg U_1\wedge U_0$. Note that since paths to a leaf of the decision tree correspond to a conjunction of decisions, sets of leaf nodes can express disjunctions of conjunctions, e.g. the set of red leaves in Figure~\ref{fig:dt} express the following disjunction of conjunctions
\[
  (\neg U_1\wedge \neg U_0) 
  \vee(\neg U_1\wedge U_0)
  \vee(U_1\wedge U_0).
\]
Since two leaves in the set have the same parent node, we can simplify the formula to
\[
\neg U_1\vee (U_1\wedge U_0)
\]
in the above example. Formally, we have the following:

\begin{definition}\label{def:dt}
  A \emph{decision tree} is a binary rooted tree $T$ in which each inner node (i.e. not a leaf) $u$ is labeled with a splitting decision $\varphi_u$. For each leaf $v$ we then define
  \[\chi_v := \bigwedge\{\varphi_u : u\in\text{path}(v)\}\]
  where $\text{path}(v)$ is the path from the root to $v$. For each set of leaves $M$ we define
  \[\chi_M := \bigvee\{\chi_v : v\in M\}.\]
  We often refer to the set of leaves as a leaf set.
\end{definition}

When learning from data, we can obtain a decision tree by iteratively partitioning the sample space. Given a feature matrix $U$ we choose a feature $j$ such that each of the partitions $\{v_i : U_{ij} = 1\}$ and $\{v_i : U_{ij} = 0\}$ have maximal homogeneity according to some measure, e.g. partitions that minimize the variance of the numerical labels. If $U$ does not only contain binary but also numerical features, in addition to the feature $j$ we determine a threshold $t$ such that the homogeneity of $\{v_i : U_{ij}\leq t\}$ and $\{v_i : U_{ij}>t\}$ is maximal. This process is then recursively repeated for each partition until a stopping criterion is met.

\begin{example}
  Let us forget for a moment the graph structure from Example~\ref{ex:graph} and consider just the feature matrix $U$, consisting of four samples $v_0, v_1, v_2, v_3$, as well as the following target values $Y$:
  \[
    U = \begin{pmatrix}
      0 & 1\\
      1 & 0\\
      0 & 0\\
      1 & 1
    \end{pmatrix}\hspace{2cm} Y = \begin{pmatrix}
      0.2\\
      0.8\\
      0.9\\
      0.1\\
    \end{pmatrix}
  \]
   Given our features $U_0$ and $U_1$, there are two possible splits, splitting on $U_0$ gives us the partition containing the sets $\{v_0, v_2\}$ and $\{v_1, v_3\}$, corresponding to $U_0$ being true and false respectively. While splitting on $U_1$ gives us a partition containing the sets $\{v_1, v_2\}$ and $\{v_0, v_3\}$, corresponding to $U_1$ being true and false respectively. Note that when splitting to minimize the variance of the $Y$ values within each partition, $U_1$ constitutes the preferable splitting criterion. 
\end{example}

\section{Iterated Decision Trees}\label{Sec:IDT}

In this section, we introduce the formal structure of our distillation model -- the \emph{Iterated Decision Tree} (IDT). IDTs consist of a sequence of decision trees. Each leaf set of a decision tree \emph{layer} in an IDT represents an $\mathcal{EMLC}$ formula with a free variable. Each subsequent decision tree layer adds a modal parameter or a new Boolean combination of leaf sets of the previous layer. Hence, a $k$ layer IDT can represent an $\mathcal{EMLC}$ formula with up to $k$ nested model parameters. In the following we formally define a single IDT layer.

\begin{definition}
\label{def: IDT_layers}
  An iterated decision tree layer $L$ consists of
  \begin{itemize}
    \item a  decision tree $T$ with splitting decisions of the form $S\varphi > n$ where $S$ is a modal parameter, and $n\in\mathbb N$,
    \item and a set of leaf sets $\{M_j\}_{j\in [l]}$ of $T$.
  \end{itemize}
\end{definition}

\begin{example}\label{ex:layer}
  Consider the following iterated decision tree layer
  
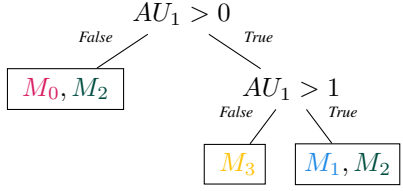
\begin{center}
    \begin{tikzpicture}[level 1/.style={sibling distance=30mm, level distance=10mm},level 2/.style={sibling distance=15mm}]
      \node {$\:AU_1 > 0$}
        child {
          node [draw] {$\:\textcolor{color1}{M_0}, \textcolor{color3}{M_2}\:$}
          edge from parent node[above left, draw=none] {\tiny False}}
        child {
          node {$\:AU_1 > 1\:$}
            child {node [draw] {$\:\textcolor{color4}{M_3}$}
            edge from parent node[above left, draw=none] {\tiny False}}
            child {node [draw] {$\textcolor{color2}{M_1}, \textcolor{color3}{M_2}$}
            edge from parent node[above right, draw=none] {\tiny True}}
        edge from parent node[above right, draw=none] {\tiny True}};
    \end{tikzpicture}
  \end{center}
where at each leaf the respective sets that contain it are indicated. That is, the left leaf appears in the leaf set $M_0$ and $M_2$, the middle leaf in only the leaf set $M_3$, and the right leaf in the leaf sets $M_1$ and $M_2$. The resulting formulas according to Definition~\ref{def:dt} are then as follows:
  \begin{align*}
    \chi_{\textcolor{color1}{M_0}} \Longleftrightarrow&\:\neg (AU_1 > 0)\\
    \chi_{\textcolor{color2}{M_1}} \Longleftrightarrow&\:(AU_1 > 0)\wedge (AU_1 > 1)\\
    \chi_{\textcolor{color3}{M_2}} \Longleftrightarrow&\:\neg (AU_1 > 0)\vee ((AU_1 > 0)\wedge (AU_1 > 1))\\
    \chi_{\textcolor{color4}{M_3}} \Longleftrightarrow
    &\:\neg (AU_1 > 0)\wedge \neg(AU_1 > 1)
  \end{align*}
  They simplify as follows:
  \begin{align*}
    \chi_{\textcolor{color1}{M_0}} \Longleftrightarrow&\:(AU_1 = 0)\\
    \chi_{\textcolor{color2}{M_1}} \Longleftrightarrow&\:(AU_1 > 1)\\
    \chi_{\textcolor{color3}{M_2}} \Longleftrightarrow&\:\neg(AU_1= 1)\\
    \chi_{\textcolor{color4}{M_3}} \Longleftrightarrow&\:(AU_1= 1)\\
  \end{align*}
\end{example}

\begin{definition}
  A sequence of $k$ iterated decision tree layers $\{L_i\}_{i\in[k]}$ is an \emph{iterated decision tree} if
    for every $i \in [k]$ and splitting decision $S\varphi >n$ occurring in $L_{i}$, $\varphi$ is of depth $0$ or $\varphi \Leftrightarrow \chi_{M}$ for some leaf set $M$ of $L_{l}$ for $l < i$. We write $M_j^i$ for the $j$-th leaf set of $L_i$ and $\chi_j^i$ for $\chi_{M_j^i}$.
\end{definition}

Hence, the $l^{th}$ IDT-layer represents $\mathcal{EMLC}$ formulas with one free variable and depth of up to $l$. Every subsequent layer can add one additional modal parameter to the formulas represented by previous IDT layers and can also create Boolean combinations of these new formulas.
\begin{example}\label{ex:idt}
  Consider an iterated decision tree consisting of two layers, the first layer as in Example~\ref{ex:layer}. Now we can add a second layer that checks if there is more than one neighbor for which $\chi_{\textcolor{color3}{M_2}} =\chi_{\textcolor{color3}{M_2^0}}$ is true.
  
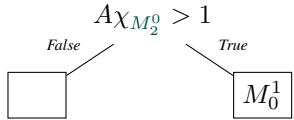
\begin{center}
    \begin{tikzpicture}[level 1/.style={sibling distance=30mm, level distance=10mm},level 2/.style={sibling distance=10mm}]
      \node {$\:A\chi_{\textcolor{color3}{M_2^0}} > 1$}
            child {node [draw] {\phantom{$M^0_1$}}
            edge from parent node[above left, draw=none] {\tiny False}}
            child {node [draw] {$M_0^1$}
            edge from parent node[above right, draw=none] {\tiny True}};
    \end{tikzpicture}
  \end{center}
  It has a single leaf set $M_0^1$ containing only the right leaf. Then
  \begin{align*}
    \chi_0^1 \Longleftrightarrow & %
    A(\neg(AU_1 = 1)) > 1 \\
  \end{align*}
Hence the combined formula labels any node one that has more than one neighbor $w$ such that $w$ does not have exactly one neighbor satisfying the node attribute $U_1$. 
\end{example}
We now show that any $\mathcal{EMLC}$ formula can be expressed as an iterated decision trees.

\begin{lemma}\label{lem:emlc-idt}
 Given a finite set of $\mathcal{EMLC}$ formulas $\{\psi_i\}_{i\in [l]}$ of depth at most $k > 0$ there is an iterated decision tree with $k$ layers such that $\chi_i^{k-1}$ is equivalent to $\psi_i$.
\end{lemma}

\def\emlcidtproof{
  \noindent Before we prove it, we show the following auxiliary result:

  \begin{lemma}\label{lem:dt}
      Let $\Phi = \{\varphi_j\}_{j\in[h]}$ be a finite set of formulas. There is a decision tree with splitting decisions in $\Phi$ such that for every Boolean combination $\psi$ of formulas in $\Phi$ there is a leaf set $M$ such that $\chi_M\Leftrightarrow\psi$.
  \end{lemma}
  \begin{proof}
      Let $T$ be the complete binary tree of depth $h$ such that each node of depth $j$ is labeled with the splitting decision $\varphi_j$. Observe that for each conjunction $c$ of the form $l_0\wedge\dots\wedge l_{h-1}$ where $l_j$ is $\neg\varphi_j$ or $\varphi_j$ there exists a leaf $t$ in $T$ such that $\chi_t = c$.
      
      Suppose $\psi$ is a Boolean combination of the $\varphi_j$. By the Disjunctive Normal Form Theorem~\citep{howson2005logic} there is a set $\mathcal C$ of conjunctions of the form $l_0\wedge\dots\wedge l_{h-1}$ where $l_j$ is $\neg\varphi_j$ or $\varphi_j$ such that \[\psi\Leftrightarrow \bigvee\mathcal C.\]
      The claim then follows by Definition~\ref{def:dt}.
  \end{proof}

  \begin{proof}[Proof of Lemma~\ref{lem:emlc-idt}]
  Note that $I\varphi > 0$ is equivalent to $\varphi$ for any formula $\varphi$. Therefore, we can rewrite any formula of depth $i$ as a formula of depth $j$ for any $j\geq i$.
  Thus, we can view any formula $\psi$ of depth $k > 0$ as a Boolean combination of formulas of the form $S\varphi > n$, where $S$ is a modal parameter and $\varphi$ is of depth $k - 1$. For instance, consider the following formula of depth $2$:
  \[(1(AU_0 = 0) > 2)\wedge(AU_1 = 1)\wedge U_0.\]
  We can rewrite it to the equivalent formula
  \[(1(AU_0 = 0) > 2)\wedge (I(AU_1 = 1)> 0) \wedge (I(IU_0> 0)> 0).\] We proceed by induction on the maximal depth of the formulas in $\{\psi_i\}_{i \in [l]}$, i.e., $k$.

  Suppose $k=1$. That is, each formula $\psi_i$ is of depth at most $1$. As explained earlier, we can always add the modal parameter $I$ to get a formula of depth 1 from a formula of depth zero. Hence, we may assume that there are formulas $\{\varphi_j\}_{j\in[h]}$ such that 
  \begin{itemize}
  \item each $\varphi_j$ is of the form $S\varphi > n$ where $\varphi$ has depth $0$
  \item and each $\psi_i$ is a Boolean combination of a subset of $\{\varphi_j\}_{j\in[h]}$.
  \end{itemize}
  By Lemma~\ref{lem:dt} there is an IDT Layer $L_0$ with splitting decisions in $\{\varphi_j\}_{j\in[l]}$ and leaf sets $M_0^0\dots M_l^0$ such that $\psi_i\Leftrightarrow\chi_i^0$ for $i\in[l]$. Thus $L_0$ gives us the desired $1$-layer IDT.

  Now suppose the claim holds for a given $k$ and the depth of each $\psi_i$ is bounded by $k+1$. Again, we may assume a set of formulas $\{\varphi_j\}_{j\in[h]}$ such that each $\psi_i$ is a Boolean combination of a subset of these formulas and each $\varphi_j$ is of the form $S\varphi > n$ where $\varphi$ has depth $k$. Applying the induction hypothesis, there is an IDT $L_0\dots L_{k-1}$ with $k$ layers and leaf sets $M_0^{k-1}\dots M_h^{k-1}$ such that $\varphi_j\Leftrightarrow \chi_0^{k-1}$. Lemma~\ref{lem:dt} gives us a decision tree with splitting decisions in $\{\varphi_j\}_{j\in[h]}$ and leaf sets $M_0^{k}\dots M_l^{k}$ such that $\chi_i^k\Leftrightarrow \psi_i$. Thus $L_0\dots L_k$ is the desired IDT.
  \end{proof}
}

\ifdefined\KR
  \emlcidtproof
\fi

\ifdefined\ICML
  \noindent A proof can be found in appendix~\ref{app:emlc-idt}.
  \edef\APPENDIX{\unexpanded\expandafter{\APPENDIX}\unexpanded{
    \section{Proof of Lemma~\ref{lem:emlc-idt}}\label{app:emlc-idt}
    Let us first recall the claim:
    \begin{manuallemma}{\ref{lem:emlc-idt}}
      Given a finite set of $\mathcal{EMLC}$ formulas $\{\psi_i\}_{i\in [l]}$ of depth at most $k > 0$ there is an iterated decision tree with $k$ layers such that $\chi_i^{k-1}$ is equivalent to $\psi_i$.
    \end{manuallemma}
    \emlcidtproof
  }}
\fi
The converse of Lemma~\ref{lem:emlc-idt} holds by definition. Hence, we have the following theorem.

\begin{theorem}
  For every $\mathcal{EMLC}$ formula $\psi$ of depth $k > 0$ there is an iterated decision tree with $k$ layers such that $\chi_0^k$ is equivalent to $\psi$. Conversely, for every iterated decision tree with $k > 0$ layers and associated formula $\chi_j^{k-1}$, there is an equivalent $\mathcal{EMLC}$ formula $\psi$ of depth $k$.
\end{theorem}

\section{Learning Iterated Decision Trees}\label{Sec:Training}

We now show how an iterated decision tree (IDT) can be learned from a given GNN. We are given a set of attributed graphs $\mathcal G$
and a $l$ layer GNN learned on $\mathcal{G}$, using the true labels of $\mathcal{G}$. We use $\mathcal{X}^{(k)}$ to denote the set of all node representations, for all the graphs in $\mathcal{G}$, computed after $k$ iterations of the GNN (see Equation~\eqref{eq: GNN_Recursion}). Note that $\mathcal{X}^{(0)}$ are simply the original node attribute matrices of all the graphs in $\mathcal{G}$. 
We define ${\mathcal{X}^{(l+1)}}$ as the graph labels returned by the GNN.
Finally, let $\mathcal{X}_{\mathrm{GNN}} = \{ \mathcal{X}^{(k)} \ | \ k \in [l+1]\}$. Algorithm~\ref{alg:LearnIDT}  assumes that a function $\mathsf{LearnIDTLayer}(\mathcal G, \mathcal U,  \mathcal{X}^{(k+1)})$ is able to learn an iterated decision tree layer given a set of graphs $\mathcal{G}$, a set $\mathcal{U}$ containing node attribute matrices for each graph in $\mathcal{G}$, and the labels $\mathcal{X}^{(l+1)}$. 
The function $\mathsf{LeafSets}$ accepts an IDT layer and returns its set of leaf sets $\{M_j\}_{j\in[l]}$ (see Definition~\ref{def: IDT_layers}).
We discuss the subroutine $\mathsf{LearnIDTLayer}$ in Section~\ref{sec:layerlearning}.

\newenvironment{ALC@prc}{\begin{ALC@g}}{\end{ALC@g}}
\begin{algorithm}
\caption{Learning Procedure for Iterated Decision Trees}
\label{alg:LearnIDT}
\begin{algorithmic}[1]
\PROCEDURE{LearnIDT}{$\mathcal G, \mathcal{X}_{\mathrm{GNN}}$}
    \STATE IDT$\gets \emptyset$
    \alglinelabel{alg:learnIDT:initiate_IDT}
    \STATE $\mathcal{U} \gets \mathcal{X}^{(0)}$
    \alglinelabel{alg:learnIDT:initiate_U}
    \FOR{$k\in[l+1]$}
    \alglinelabel{alg:LearnIDT:loop}
        \STATE $L\gets \mathsf{LearnIDTLayer}(\mathcal G, \mathcal U, \mathcal{X}^{(k+1)})$
        \alglinelabel{alg:LearnIDT:learnlayer}
        \FOR{$M\in\mathsf{LeafSets}(L)$}
        \alglinelabel{alg:LearnIDT:listleafsets}
            \STATE $\mathcal{U}\gets \mathsf{Append}(\mathcal U, \mathsf{FeatureVector}(\mathcal G, \chi_M))$
            \alglinelabel{alg:LearnIDT:addfeature}
        \ENDFOR
        \STATE IDT$\gets \mathsf{Append}($IDT, $L)$
        \alglinelabel{alg:LearnIDT:addlayer}
    \ENDFOR \alglinelabel{alg:LearnIDT:addlayer:Endfor}
    \STATE \textbf{return} IDT
\ENDPROCEDURE
\end{algorithmic}
\end{algorithm}

In Algorithm~\ref{alg:LearnIDT}, we initialize an empty IDT and the set $\mathcal U$ with the original node attribute matrices. 
Each iteration of the main loop (Line~\ref{alg:LearnIDT:loop}--\ref{alg:LearnIDT:addlayer:Endfor}) computes a new IDT layer $L$ ( Line~\ref{alg:LearnIDT:learnlayer}), using the graphs $\mathcal{G}$ with updated node-attribute matrices $\mathcal U$, and target labels $\mathcal{X}^{(k+1)}$ obtained from the GNN. 
Each leaf set $M$ identified by $\mathsf{LeafSets}$ (Line~\ref{alg:LearnIDT:listleafsets}) corresponds to a disjunction of conjunctions $\chi_M$ (cf. Definition~\ref{def:dt}).
We evaluate $\chi_M$ for every node in $\mathcal{G}$ and add an explicit feature vector to the node attribute matrices in $\mathcal{U}$ (Line~\ref{alg:LearnIDT:addfeature}).
The new set of node-attribute matrices $\mathcal{U}$ is then used as the node attributes matrix in the subsequent loop iteration.

\begin{example}
Recall the graph from Example~\ref{ex:graph}. Assume that $\mathcal{G}$ consists of only this graph and graph label. Suppose that $\mathsf{LearnIDTLayer}(\mathcal G, \mathcal{U},\mathcal{X}_{0})$ yields the IDT layer shown in Example~\ref{ex:layer}.
  Then we extend the node attributes with binary vectors representing  $\chi_M$ for each leaf set $M \in \{ \textcolor{color1}{M_0}, \textcolor{color2}{M_1}, \textcolor{color3}{M_2}, \textcolor{color4}{M_3} \}$. For $\textcolor{color1}{M_0}$ we add
  \begin{align*}
     (AU_1 = 0) &= (A(1\ 0\ 0\ 1)^T = 0)\\ &= ((0\ 2\ 1\ 0)^T= 0)\\ &= {\color{color1}(1\ 0\ 0\ 1)^T}
  \end{align*}
  For $\textcolor{color2}{M_1}, \textcolor{color3}{M_2}, \textcolor{color4}{M_3}$ we add
  \begin{align*}
    (AU_1 > 1) &= {\color{color2}(0\ 1\ 0\ 0)^T}\\
    \neg(AU_1 = 1)& = {\color{color3}(1\ 1\ 0\ 1)^T}\\
    (AU_1 = 1) &= {\color{color4}(0\ 0\ 1\ 0)^T}
  \end{align*}
  respectively. The result is an extended node attribute matrix
  \[\mathcal{U} = \begin{pmatrix}
    0 & 1 &\color{color1}1&\color{color2}0&\color{color3}1&\color{color4}0\\
    1 & 0&\color{color1}0&\color{color2}1&\color{color3}1&\color{color4}0\\
    0 & 0&\color{color1}0&\color{color2}0&\color{color3}0&\color{color4}1\\
    1 & 1&\color{color1}1&\color{color2}0&\color{color3}1&\color{color4}0
   \end{pmatrix}
\]
\noindent
which is then used for learning the next IDT layer.
\end{example}

\subsection{Learning Iterated Decision Tree Layers}
\label{sec:layerlearning}

We will now describe the procedure $\mathsf{LearnIDTLayer}$. It consists of two subprocedures, learning the underlying decision tree and choosing the leaf sets.
\subsubsection{Obtaining the Decision Tree}
Given a set of graphs $\mathcal G$, a set of node attribute matrices $\mathcal U$ and node representations $\mathcal X^{(k+1)}$, we organize them into a table as illustrated in Figure~\ref{fig:table}:
\begin{itemize}
    \item There is a row for each node in the dataset.
    \item For each modal parameter $S$ and considered feature $U_j$, there is one column labeled $SU_j$. Additionally, there is a final column for the GNN node representations $\mathcal X^{(k+1)}$.
    \item The entry at the intersection of the row associated with a vertex $v$ and the column labelled with $SU_j$ contains the
    number of nodes $w\in \varepsilon_S(v)$ such that $U_j(w)$ holds.
    \item The entries in the final column are the node representations $\mathcal X^{(k+1)}$ which are prediction targets for learning the decision tree.
\end{itemize}

We can learn a decision tree from this data using conventional decision tree learning algorithms such as \citet{scikit-learn}. Note that splitting on the first column associated with modal parameter $S$ and feature $U_j$ with threshold $n$ corresponds to splitting on the $\mathcal{EMLC}$ formula $SU_j > n$.

\subsubsection{Obtaining the Leaf Sets}

Finally, to obtain an IDT layer, we need to determine the leaf sets. If we consider all $2^k$ possible sets of leaves, this leads to an explosion of the number of node attributes. We have observed that most formulas obtained in this manner are also not used in the subsequent layers. Hence, we propose a heuristic:

Since the decision tree was learned with numeric prediction targets, each leaf of the decision tree is associated with a numeric prediction.
We group leaves with similar predictions since they are represented similarly by the GNN. To this end, we perform agglomerative, hierarchical clustering~\citep{gan2020data}. We start with all leaf sets containing a single leaf. Then we iteratively merge the pair of leaf sets with the closest numerical values until one set remains. This approach yields $2k-1$ instead of $2^k$ leaf sets.

Our approach allows us to go from the formulas associated with singleton leaf sets to formulas associated with larger leaf sets, and finally the formula associated with the set of all leaves, i.e. $\top$ which is always true. Adding $\top$  as a feature allows representing the degree of a node in the next layer, since $(G, v)\models A\top > n$ if and only if $d_v > n$.

\begin{figure}[t]
    \centering
    \begin{tabular}{c|ccccc}
         &&$IU_0$ &$AU_0$ &$\cdots$ & $\mathcal X^{(k+1)}$\\\hline
         $v_0$& $\cdots$&$0$&$4$ & $\cdots$ & $(0.32, \dots, 0.82)^T$\\
         $v_1$& $\cdots$&$1$&$1$ &$\cdots$ & $(0.92, \dots, 0.12)^T$\\
         \vdots&&\vdots& $\vdots$ & &$\vdots$\\
         $v_n$ &  $\cdots$ & $0$  & $2$  &$\cdots$ & $(0.24, \dots, 0.55)^T$
    \end{tabular}
    \caption{Schematic representation of the data table.}
    \label{fig:table}
\end{figure}

\subsection{Practical Considerations}

We have made a number of choices to arrive at a practical implementation of our proposed algorithm.
\begin{enumerate}
    \item For all but the final iterated decision tree layer, we consider only the modal parameters $I, A, I + A$, which correspond to modal parameters quantifying only on the local neighborhood. As the aggregation operation of the GNN does not have access to features of non-neighbors, local quantification captures operations performed by real-world GNNs. Limiting ourselves to these modal parameters did not make a notable difference in the model performance and significantly reduced the computational cost.
    \item As our experiments consist of graph classification tasks, we only consider the modal parameter $1$ for the final iterated decision tree layer. A formula with an outer-most modal parameter $1$ provides identical labels for all nodes. Such a formula can thus be viewed as a graph classifier.
    \item We limit the depth of all but the final iterated decision tree layers to two to enhance interpretability. For the final IDT layer, we don't limit the depth but instead perform minimal cost complexity pruning~\citep{breiman2017classification}.
    \item Instead of a single decision tree, we train multiple decision trees with a randomized subset of the features at each layer and add the formulas corresponding to each of their leaf sets. Empirically, this leads to better generalization.
\end{enumerate}

\section{Relative Modal Parameters}
\label{sec:cp}

In GNN architectures, such as the GCN, the aggregation is normalized by the degree of the node. However, this normalization is not expressible in $\mathcal{EMLC}$ and can, as such, not be captured by the defined iterated decision tree. We argue that a property like
\emph{``more than half of the neighbors of $v$ satisfy $U_0$''}
should be expressible in our model. Therefore, we propose the following extension of $\mathcal{EMLC}$:

\begin{definition}
\label{def: EMLC_2}
   An $\mathcal{EMLC}\%$ formula is defined by the following grammar:
  \[\varphi ::= U_j \mid \top\mid\varphi\wedge\varphi \mid \varphi\vee\varphi \mid\neg\varphi \mid S\varphi > n \mid S\varphi > p\]
  where $S$ ranges over modal parameters (cf. Definition~\ref{def: EMLC}), $n$ over $\mathbb N$ and $p$ over the open interval $(0, 1)$. The semantics of the first six cases agree with $\mathcal{EMLC}$, while ${(G, v)\models S\varphi > p}$ holds if and only if there are more than $p\cdot |\varepsilon_S(v)|$ vertices $w\in \varepsilon_S(v)$ such that $(G, w)\models S\varphi$.
  Note that we can always distinguish between $S\varphi > n$ and $S\varphi > p$ , since $n$ and $p$ range over disjoint sets.
\end{definition}

\begin{example}\label{ex:EMLCp}
  Using the semantics defined in Definition~\ref{def: EMLC_2}, the formula \[1U_0 > 0.5\] holds if more than half the nodes in $G$ satisfy $U_0$.
\end{example}

\begin{theorem}
  $\mathcal{EMLC}\%$ is more expressive than $\mathcal{EMLC}$.
\end{theorem}

\begin{proof}
Using Theorem~\ref{thm:c2-emlc} we have that every $\mathcal{EMLC}$ formula is equivalent to a $\C2$ formula. Furthermore, every $\C2$ formula can be captured in first-order logic \citep{grohe2021logic}. We will now show that there is an $\mathcal{EMLC\%}$ formula that can not be expressed in first-order logic. Our proof relies on zero-one laws on images as proved in \citet{coupier2004zero}.

    Consider graphs with only one unary node attribute $U_0$. Assume for each node $v$ that $U(v)$ is true with a probability $0.5$. Define $\varphi := IU_0 > 0.5.$ The probability of $\varphi$ holding in a graph converges to exactly 0.5 as the graph cardinality grows. Hence, $\varphi$ does not obey a zero-one law. Therefore, $\varphi$ is not definable in first-order logic.
\end{proof}

The following result transfers from~\citet[Theorem 5.2]{Barcel2019LogicalEO} by adapting case 4 of their proof to relative modal parameters, which can be easily done by adding comparisons of the form $S\varphi > p$:
\begin{theorem}
  $\mathcal{EMLC}\%$ formulas can be computed by GNNs.
\end{theorem}

However, the property of a node having more red than blue neighbors is still not expressible in $\mathcal{EMLC}\%$. Whereas, this property can be computed by a GNN \citep{Grohe2023TheDC}. This puts the expressivity of $\mathcal{EMLC}\%$ strictly between the expressivity of $\C2$ and GNNs.

It is straightforward to extend IDTs to $\mathcal{EMCL}\%$ since the computational difference is a single division operation.

\section{Experiments}\label{Sec:Experiments}

\def\accuracytable{
  \begin{tabular}{c|cc|ccccc}
    Test & \multirow{2}*{GCN} & \multirow{2}*{GIN} & IDT & IDT & IDT & IDT & IDT\\
    Accuracy&&&(GCN)& (GCN+True) & (GIN)&(GIN+True)& (True)\\\hline
    AIDS & $0.92\pm 0.02$ & $0.92\pm 0.03$ & $0.99\pm 0.01$ & \colorbox{black!20!white}{$1.00\pm 0.02$}& $0.98\pm0.01$&\colorbox{black!20!white}{$1.00\pm0.00$}&\colorbox{black!20!white}{$1.00\pm0.00$}\\BZR&$0.81\pm0.06$&$0.80\pm0.08$&$0.79\pm0.08$&\colorbox{black!20!white}{$0.83\pm0.06$}&$0.80\pm0.09$&\colorbox{black!20!white}{$0.83\pm0.06$}&$0.81\pm0.04$\\
    PROTEINS & $0.72\pm 0.04$ & $0.73\pm 0.04$ &$ 0.73\pm0.04$&\colorbox{black!20!white}{$0.74\pm0.03$}&$0.73\pm0.03$&$0.72\pm0.02$&$0.71\pm0.03$\\
    $\psi_0$ & \colorbox{black!20!white}{$1.00\pm 0.00$} & \colorbox{black!20!white}{$1.00\pm 0.00$} & \colorbox{black!20!white}{$1.00\pm 0.00$} & \colorbox{black!20!white}{$1.00\pm 0.00$}& \colorbox{black!20!white}{$1.00\pm 0.00$}& \colorbox{black!20!white}{$1.00\pm 0.00$} & \colorbox{black!20!white}{$1.00\pm 0.00$}\\
    $\psi_1$ & $0.88\pm 0.02$ & $0.88\pm 0.02$ & $0.93\pm0.02$&\colorbox{black!20!white}{$0.97\pm0.07$}&$0.92\pm0.04$&$0.95\pm0.07$&$0.96\pm0.04$\\
    $\psi_2$&$0.81\pm0.02$&$0.82\pm0.01$&$0.94\pm0.01$&$0.96\pm0.01$& $0.95\pm0.01$&$0.94\pm0.05$&\colorbox{black!20!white}{$0.99\pm0.03$}\\
    BAMulti & $0.99\pm 0.02$ & $0.97\pm 0.03$ & \colorbox{black!20!white}{$1.00\pm0.01$}&\colorbox{black!20!white}{$1.00\pm 0.00$}&$0.99\pm0.01$&\colorbox{black!20!white}{$1.00\pm0.00$}&\colorbox{black!20!white}{$1.00\pm0.00$}
  \end{tabular}
}

\def\fonetable{
  \begin{tabular}{c|cc|ccccc}
    F1-Score & \multirow{2}*{GCN} & \multirow{2}*{GIN} & IDT & IDT & IDT & IDT & IDT\\
    (macro)&&&(GCN)& (GCN+True) & (GIN)&(GIN+True)& (True)\\\hline
    AIDS & $0.88\pm0.04$ & $0.87\pm 0.03$ &$0.98\pm0.02$&\colorbox{black!20!white}{$1.00\pm0.00$}&$0.97\pm0.02$&\colorbox{black!20!white}{$1.00\pm0.00$}&\colorbox{black!20!white}{$1.00\pm0.00$}\\
    BZR&\colorbox{black!20!white}{$0.73\pm0.07$}&$0.72\pm0.10$&$0.65\pm0.12$&$0.63\pm0.08$&$0.67\pm0.13$&$0.64\pm0.09$&$0.68\pm0.05$\\
    PROTEINS & $0.71\pm 0.04$ & $0.72\pm 0.04$ &$ 0.72\pm0.04$&\colorbox{black!20!white}{$0.73\pm0.03$}&$0.72\pm0.04$&$0.70\pm0.02$&$0.69\pm0.04$\\
    $\psi_0$ & \colorbox{black!20!white}{$1.00\pm 0.00$} & \colorbox{black!20!white}{$1.00\pm 0.00$} & \colorbox{black!20!white}{$1.00\pm 0.00$} & \colorbox{black!20!white}{$1.00\pm 0.00$}& \colorbox{black!20!white}{$1.00\pm 0.00$}& \colorbox{black!20!white}{$1.00\pm 0.00$} & \colorbox{black!20!white}{$1.00\pm 0.00$}\\
    $\psi_1$ & $0.86\pm 0.03$ & $0.86\pm 0.02$ & $0.92\pm0.02$&\colorbox{black!20!white}{$0.96\pm0.09$}&$0.91\pm0.04$&$0.94\pm0.09$&$0.95\pm0.05$\\
    $\psi_2$&$0.80\pm0.02$&$0.81\pm0.01$&$0.94\pm0.01$&$0.95\pm0.01$&$0.94\pm0.01$&$0.94\pm0.05$&\colorbox{black!20!white}{$0.99\pm0.03$}\\
    BAMulti & $0.99\pm 0.02$ & $0.97\pm 0.03$ & \colorbox{black!20!white}{$1.00\pm0.01$}&\colorbox{black!20!white}{$1.00\pm0.00$}&$0.99\pm0.01$&\colorbox{black!20!white}{$1.00\pm0.00$}&\colorbox{black!20!white}{$1.00\pm0.01$}
  \end{tabular}
}

\def\fidelitygcntable{
  \begin{tabular}{c|c|cc}
    Fidelity &  \multirow{2}*{GIN} & IDT & IDT\\
    (GCN)&&(GCN)& (True)\\\hline
    AIDS &{$0.92\pm0.02$}&{$0.92\pm0.02$}&{$0.92\pm0.02$}\\
    BZR &{$0.90\pm0.05$}&$0.80\pm0.06$&$0.79\pm0.05$\\
    PROTEINS & {$0.90\pm 0.05$} &$ 0.84\pm0.04$&$0.80\pm0.06$\\
    $\psi_0$ & {$1.00\pm 0.00$} & {$1.00\pm 0.00$} & {$1.00\pm 0.00$}\\
    $\psi_1$ & {$0.94\pm 0.01$} & $0.92\pm0.01$&$0.85\pm0.02$\\
    $\psi_2$&{$0.86\pm0.01$}&$0.83\pm0.02$&$0.81\pm0.02$\\
    BAMulti & $0.97\pm 0.02$ & {$0.99\pm0.02$}&$0.98\pm0.02$
  \end{tabular}
}

\def\fidelitygintable{
  \begin{tabular}{c|c|cc}
    Fidelity & \multirow{2}*{GCN} & IDT & IDT\\
    (GIN)&&(GIN)&(True)\\\hline
    AIDS & {$0.92\pm0.02$}&$0.91\pm0.03$&$0.91\pm0.02$\\
    BZR &{$0.90\pm0.05$} & $0.80\pm0.07$&$0.77\pm0.05$\\
    PROTEINS & {$0.90\pm 0.05$} &$0.87\pm0.03$&$0.80\pm0.04$\\
    $\psi_0$ & {$1.00\pm 0.00$} & {$1.00\pm 0.00$}&{$1.00\pm 0.00$}\\
    $\psi_1$ & {$0.94\pm 0.01$} &$0.91\pm0.03$&$0.85\pm0.03$\\
    $\psi_2$&$0.86\pm0.01$&$0.82\pm0.02$&{$0.82\pm0.01$}\\
    BAMulti & {$0.97\pm 0.02$} &{$0.97\pm0.03$}&{$0.97\pm0.03$}
  \end{tabular}
}

\def\experimentslong{
  There are seven models we consider for experiments%
  \footnote{Available at \href{https://github.com/lexpk/LogicalDistillationOfGNNs}{github.com/lexpk/LogicalDistillationOfGNNs}.}:
  \begin{itemize}
    \item We evaluate two GNN architectures, GCN+GraphNorm and GIN+GraphNorm as described in~\citet{cai2021graphnorm}, simply called GCN and GIN from here on. The number of layers, hidden dimensions, and the learning rate are determined experimentally.
    \item Two IDTs as described in Section~\ref{Sec:Training}. One leveraging the GCN node representations, IDT(GCN), and the other one leveraging the GIN node representations, IDT(GIN).
    \item Two IDTs as above. However, the final layer of each IDT is learned using the true labels of the dataset instead of the GNN outputs, denoted as IDT(GCN+True) and IDT(GIN+True). Using the true labels for the final layer allows for increased accuracy while still leveraging the information of the underlying GNN.
    \item As a baseline we consider an IDT in which every layer is learned with the true labels of the dataset. This model operates purely on the data and is called IDT(True).
  \end{itemize}

  \subsection{Datasets}

    \noindent\textbf{Real world graph classification datasets} are obtained from the TU Dortmund collection \citep{Morris2020TUDataset}. They are commonly used in the GNN literature.
    \begin{itemize}
      \item AIDS~\citep{riesen2008iam} contains 2000 graphs representing molecular compounds. The label represents activity against HIV.
      \item BZR~\citep{sutherland2003spline} contains 405 graphs representing ligands for the benzodiazepine receptor. The label represents whether a threshold measuring binding affinity is crossed.
      \item PROTEINS~\citep{borgwardt2005protein} contains 1113 graphs representing proteins. The label represents whether a given protein is an enzyme or not.
    \end{itemize}
    
    \noindent\textbf{Synthetic datasets} are based on $\mathcal{EMLC}$. We generate 1000 Erd\H{o}s-R\'enyi graphs with $n=13$ and $p=0.5$ and add two node features. The feature $U_0$ is always $1$ and the feature $U_1$ is $1$ with probability $0.5$ for each node. Then, we label the graphs according to an $\mathcal{EMLC}\%$ formula. The following formulas of increasing complexity are considered:
    \begin{itemize}
      \item $\psi_0 := 1U_1 > 0.5$.\\``More than half of the nodes satisfy $U_1$.''
      \item $\psi_1 := 1((AU_0 < 4) \vee (AU_0 > 9)) > 0$.\\``There is a node $v$ such that $d_v < 4$ or $d_v > 9$.''
      \item $\psi_2 := 1(A(AU_0 > 6) > 0.5) > 0.5$\\
      ``For at least half the nodes at least half of their neighbors have degree greater than $6$''
    \end{itemize}
    
    \noindent\textbf{BAMultiShapes}~\citep{azzolin2022global} is a synthetic dataset based on subgraph motifs. The samples are generated from Barab\'asi-Albert graphs. Each node has a feature $U_0$ which is always $1$. To each graph, either nothing, a wheel graph, a house graph, or a grid graph is attached with a single edge, or a combination of these shapes. Class zero consists of all graphs with exactly two such shapes added. Class one of graphs with zero, one, or all three shapes added. Generally, the existence of such shapes is not expressible in $\mathcal{EMLC}\%$.

  \subsection{Metrics}

  We report the mean and standard deviation over a 10-fold cross-validation. 
  The same splits are used for all models. 
  The training is inductive, i.e., the test set is completely separated from the training process.
  We use three metrics for evaluation:

  \begin{itemize}
    \item The accuracy of the model.  
    \item Macro F1-score: The F1-score of each class is the harmonic mean of precision and recall. These scores are then averaged, resulting is a suitable metric for more imbalanced datasets, such as AIDS.
    \item Fidelity with regard to each GNN. This is the proportion of predictions where the model and the GNN agree.
  \end{itemize}

  \subsection{Quantitative Results}

  IDTs are able to match and beat the performance of the GNNs.
  Table~\ref{tab:accuracy} shows that IDTs trained on GCN representations and true labels consistently achieve the highest accuracy, followed by IDTs trained on GIN representations and true labels.
  This ranking remains the same under the F1-score as shown in Table~\ref{tab:f1}.
  While the performance benefits are consistent, training on the true labels alone gives only slightly worse results, suggesting that there could be merit to IDTs as a stand-alone model.

  Table~\ref{tab:fidelityGCN} and Table~\ref{tab:fidelityGIN} show the fidelity to the GCN and GIN model, respectively. Using the GNN activations generally results in IDT outputs which are closer to the GNN model, than just using the training labels. However, the two GNNs generally have larger fidelity between each other then when compared to the IDTs. This suggests that the IDTs operate in a fundamentally different way than GNNs.
  This point is further reinforced by the fact that on synthetic datasets the IDTs outperform both the GNNs.

  \subsection{Qualitative Results}\label{Sec:Explainability}

  We will now look at the interpretability of the distilled IDTs. First, we discard all decision tree layers and node sets which are not used for the final prediction. This procedure is automated and results in an equivalent, more compact IDT. %

  \subsubsection{AIDS}
  Figure~\ref{fig:explAIDS} shows an IDT for the AIDS dataset. The IDT is remarkably small. The decision tree in the first layer does not have a split condition. Hence $M^0_0 = \top$ and $\chi_0^{0}$ is true for every node. The second layer, therefore, simply computes if there are more than twelve nodes in the graph.

  \subsubsection{BAMultiShapes}
  Figure~\ref{fig:explBAMulti} shows one extracted IDT for the BAMultiShapes dataset that achieves an accuracy of 1.0 on the test set. After some calculation, we have
  \begin{align*}
      \chi_0^0 \Longleftrightarrow 
      & %
      \:AU_0 < 2\\
      \chi_0^1 \Longleftrightarrow &\:(AU_0 > 2)\wedge (AU_0 < 12)
  \end{align*}

  A graph is then classified as class 0, i.e., having exactly two shapes, if 
  \[(1\chi_0^1 > 0.312) \wedge (1\chi_0^0 > 14)\]
  is satisfied.
  Recall that $U_0$ is true for all vertices in BAMultiShapes.
  Hence, if at least $31.2\%$ of all nodes in a graph have degree between three and eleven and at least fifteen nodes have degree zero or one, a graph is assigned to Class 0.
  Due to the construction of the dataset, these observations about the degree distribution correlate strongly with the label.

\ifdefined\KR
\showthequalitativeexamples
\fi

  \paragraph{Other Datasets}
  For the other real-world datasets BZR and PROTEINS, the IDTs are more complex, often containing more than 10 decision trees. 
  Still, by carefully examining the trees it is possible to deduce explainable logical information which we leave for future work.
  For the synthetic datasets labeled with $\mathcal{EMLC}$ formulas $\psi_0$ and $\psi_1$, we recover the ground truth, i.e., an IDT equivalent to $\psi_0$ and $\psi_1$ respectively. For the deeper formula $\psi_2$ we are not able to recover the ground truth formula. The IDTs identify a more complex formula that approximates $\psi_2$.
}

\def\experimentsshort{
We have implemented IDTs and briefly highlight our empirical results. A detailed analysis of the experiments can be found in \cref{app:Extended Experiments}. Our code is available \href{https://github.com/lexpk/LogicalDistillationOfGNNs}{online}.

We measure accuracy, F1-score, and fidelity of Graph Convolutional Networks (GCNs), Graph Isomorphism Networks (GINs), and Iterative Decision Tree (IDT) variants on several real world and synthetic datasets. 
IDTs achieve high accuracy and F1-scores, outperforming both GCN and GIN models in many cases.  
Further analysis of fidelity indicates that the outputs of different GNNs are closer to each other than the outputs of a GNN and its corresponding IDT, indicating that IDTs function differently from GNNs. 
This is further supported by the superior performance of IDTs on synthetic datasets where the ground truth is an $\mathcal{EMLC}\%$ formula.

\showthequalitativeexamples

A qualitative evaluation of the IDTs demonstrates their interpretability. 
The IDT for the AIDS dataset, shown in \cref{fig:explAIDS}, simplifies the decision-making process to a rule based on the number of nodes in a graph. 
If a graph has more than twelve nodes, it is assigned to Class 0.
Similarly, the IDT for the BAMultiShapes dataset, shown in \cref{fig:explBAMulti} effectively uses degree-based features to classify graphs.
If at least $31.2\%$ of all nodes in a graph have degree between three and eleven and at least fifteen nodes have degree zero or one, a graph is assigned to Class 0.
This showcases the model's ability to derive meaningful, simple decision rules.

}

\def\showthequalitativeexamples{
\begin{figure}
    \begin{center}
      \begin{tikzpicture}[level 1/.style={sibling distance=30mm, level distance=10mm},level 2/.style={sibling distance=10mm}]
          \node[rectangle, draw=black] at (0,0) {$M^{0}_0$};
          
            \node[rectangle] at (0mm,-10mm) {$1 \chi^{0}_{0} > 12$}
              child {
                node[rectangle, draw=black] {Class 1} {
                } edge from parent node[above left,draw=none] {\tiny False}
              }
              child {
                node[rectangle, draw=black] {Class 0} {
                }  edge from parent node[above right,draw=none] {\tiny True}
              };

      \draw[dashed] (-40mm,-5mm) -- (40mm,-5mm);
      \node[anchor=south west] at (-40mm,-5mm) {\footnotesize Layer 0};
      \node[anchor=north west] at (-40mm,-5mm) {\footnotesize Layer 1};
      
      \end{tikzpicture}
      \caption{Distilled IDT for AIDS. At layer 0, we have just one leaf set $M^{0}_0$ containing only the tree root. Hence, the formula  $\chi_0^{0}$ is equivalent to $\top$. The rule derived for class 0 is thus $1\top > 12$. It expresses that the graph has more than $12$ nodes.}
      \label{fig:explAIDS}
      \end{center}
\end{figure}
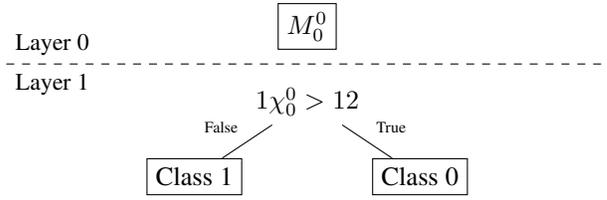
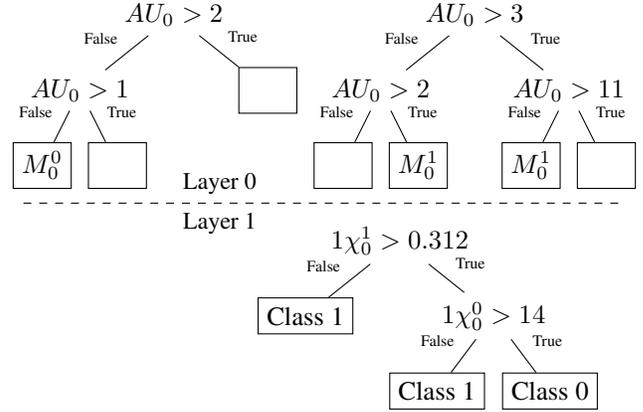
\begin{figure}[t]
    \begin{center}
      \vspace{.2cm}

      \begin{tikzpicture}[level 1/.style={sibling distance=25mm, level distance=10mm},level 2/.style={sibling distance=10mm}]
      
        \node[rectangle] at (0,0) {$A U_0 > 2$}
          child {
            node[rectangle] {$A U_0 > 1$} {
              child {node[rectangle, draw=black] {$M_0^0$} edge from parent node[above left, draw=none] {\tiny False}}
              child {node[rectangle, draw=black] {\color{white}$M^1_0$} edge from parent node[above right, draw=none] {\tiny True}}
            } edge from parent node[above left,draw=none] {\tiny False}
          }
          child {
            node[rectangle, draw=black] {\color{white}$M^1_0$} {
            }  edge from parent node[above right,draw=none] {\tiny True}
          };

          \node[rectangle] at (40mm,0) {$A U_0 > 3$}
          child {
            node[rectangle] {$A U_0 > 2$} {
              child {node[rectangle, draw=black] {\color{white}$M^1_0$} edge from parent node[above left, draw=none] {\tiny False}}
              child {node[rectangle, draw=black] {$M^1_0$} edge from parent node[above right, draw=none] {\tiny True}}
            } edge from parent node[above left,draw=none] {\tiny False}
          }
          child {
            node[rectangle] {$A U_0 > 11$} {
              child {node[rectangle, draw=black] {$M^1_0$} edge from parent node[above left, draw=none] {\tiny False}}
              child {node[rectangle, draw=black] {\color{white}$M^1_0$} edge from parent node[above right, draw=none] {\tiny True}}
            }  edge from parent node[above right,draw=none] {\tiny True}
          };

            \node[rectangle] at (30mm,-30mm) {$1 \chi_0^1 > 0.312$}
          child {
            node[rectangle, draw=black] {Class 1} {
            } edge from parent node[above left,draw=none] {\tiny False}
          }
          child[level 2/.style={sibling distance=15mm}] {
            node[rectangle] {$1 \chi_0^0 > 14$} {
              child {node[rectangle, draw=black] {Class 1} edge from parent node[above left, draw=none] {\tiny False}}
              child {node[rectangle, draw=black] {Class 0} edge from parent node[above right, draw=none] {\tiny True}}
            }  edge from parent node[above right,draw=none] {\tiny True}
          };

      \draw[dashed] (-20mm,-25mm) -- (60mm,-25mm);
      \node[anchor=south west] at (-0mm,-25mm) {\footnotesize Layer 0};
      \node[anchor=north west] at (-0mm,-25mm) {\footnotesize Layer 1};
          
      \end{tikzpicture}
      \caption{Distilled IDT for BAMultiShapes. $\chi_0^0\Leftrightarrow AU_0 < 2$ and $\chi_0^1\Leftrightarrow (AU_0 > 2)\wedge (AU_0< 12)$. The rule derived for class $0$ is $(1((AU_0 > 2)\wedge (AU_0< 12)) > 0.312)\wedge (1(AU_0 < 2) > 14).$ It expresses that for more than $31.2\%$ of nodes their degree $d_v$ satisfies $2< d_v< 12$ and for more than $14$ nodes $d_v < 2$.}
      \label{fig:explBAMulti}
      
    \end{center}
  \end{figure}
}

\ifdefined\KR
  \fboxsep1pt
  \begin{table*}[t]
  \begin{center}
  \accuracytable
  \caption{Accuracy of our proposed IDT method and of GNN models we distill from.}
  \label{tab:accuracy}
  \vspace{0.4cm}
  \end{center}
  \begin{center}
  \fonetable
  \caption{F1-Score with macro aggregation.}
  \label{tab:f1}
  \vspace{0.4cm}
  \end{center}

  \begin{minipage}[t]{0.5\textwidth}
  \fidelitygcntable
  \caption{Fidelity with regard to the predictions of the GCN.}
  \label{tab:fidelityGCN}
  \end{minipage}
  \begin{minipage}[t]{0.5\textwidth}
      
  \fidelitygintable
  \caption{Fidelity with regard to the predictions of the GIN.}
  \label{tab:fidelityGIN}
  \end{minipage}
  \vspace{-.3cm}
  \end{table*}

  \experimentslong
\fi

\ifdefined\ICML
  \experimentsshort
  \edef\APPENDIX{\unexpanded\expandafter{\APPENDIX}\unexpanded{
    \section{Extended Experiments}\label{app:Extended Experiments}
  \fboxsep1pt
  \begin{table*}[t]
  \begin{center}
    \vskip -0.2in
    \caption{Accuracy of our proposed IDT method and of GNN models we distill from.}
    \label{tab:accuracy}
    \accuracytable
  \vspace{0.4cm}
  \end{center}
  \begin{center}
      \vskip -0.3in
      \caption{F1-Score with macro aggregation.}
      \label{tab:f1}
      \fonetable
  \vspace{0.4cm}
  \end{center}

  \begin{minipage}[t]{0.5\textwidth}
      \vskip -0.4in
      \caption{Fidelity with regard to the predictions of the GCN.}
  \label{tab:fidelityGCN}
      \vskip 0.1in
      \fidelitygcntable
  \end{minipage}
  \begin{minipage}[t]{0.5\textwidth}
      \vskip -0.4in
      \caption{Fidelity with regard to the predictions of the GIN.}
    \label{tab:fidelityGIN}
      \vskip 0.1in
      \fidelitygintable
  \end{minipage}
  \vspace{-.3cm}
  \end{table*}

  \experimentslong
  }}
\fi

\section{Conclusion}\label{Sec:Conclusion}
We have presented Iterated Decision Trees (IDTs), a novel model tailored towards distilling interpretable logical formulas from  Graph Neural Networks (GNNs). IDTs can express any logical formula expressible in first order logic with two variables and counting quantifiers ($\C2$) --- a fragment of first-order logic that is closely connected to the logical expressivity of GNNs. We have also introduced an extension of $\C2$ that captures operations commonly used in GNNs, such as mean aggregation. This extension was easily incorporated into IDTs without any significant computational overhead. The distilled IDTs often surpass the accuracy of the underlying GNN while providing insight into the decision process. 
They also outperform the considered GNNs when the ground truth is itself a logical formula.
The classification decisions of the IDT are interpretable and enable us to extract insights on multiple datasets.

\section{Future Work and Limitations}\label{Sec:Future Work}
In this work, we have applied IDTs only to simple undirected graphs. Loops and multi-edges, however, can be incorporated into our model by allowing entries on the diagonal of the adjacency matrix $A$ and allowing integer entries respectively. Generalizing to directed graphs is also possible. In this case the adjacency matrix is no longer symmetric, so the modal parameter $A$ only represents out-neighbors. Thus, we need to introduce a new modal parameter $A^T$ to represent in-neighbors and its combinations with the other modal parameters. Extending the proposed method to multi-relational graphs and edge labels would require further changes, which we will consider in future work.

While our approach has allowed us insights into the AIDS and BAMultiShapes datasets, we found it more difficult to extract meaningful explanations for other datasets. We plan to further analyze the obtained explanations and apply regularization techniques in order to obtain more human-readable explanations.
As IDTs perform reasonably even without GNNs, we also look forward to further assessing the merit of IDTs as an independent architecture. %

\section*{Acknowledgments}

This work was funded in part by the Vienna Science and Technology Fund (WWTF), project StruDL (ICT22-059); by the Austrian Science Fund (FWF), project NanOX-ML (6728); and by the European Unions Horizon Europe Doctoral Network programme under the Marie-Skłodowska-Curie grant, project Training Alliance for Computational systems chemistry (101072930). SM would like to thank Steve Azzolin for the fruitful discussions.

\end{document}